# Efficient Search-Based Inference for Noisy-OR Belief Networks: TopEpsilon

**Kurt Huang**
Section on Medical Informatics
Medical School Office Building X-215
Stanford, CA 94305
<khuang@camis.stanford.edu>

**Max Henrion**
Institute for Decision Systems Research
4984 El Camino Real, Suite 110
Los Altos, CA 94022
<henrion@lumina.com>

## Abstract

Inference algorithms for arbitrary belief networks are impractical for large, complex belief networks. Inference algorithms for specialized classes of belief networks have been shown to be more efficient. In this paper, we present a search-based algorithm for approximate inference on arbitrary, noisy-OR belief networks, generalizing earlier work on search-based inference for two-level, noisy-OR belief networks. Initial experimental results appear promising.

## 1  INTRODUCTION

The computational complexity of exact inference on Bayesian networks is NP-hard (Cooper, 1990). For small networks, inference is still practical. However, for large, richly-connected networks such as QMR-BN (Shwe et al., 1991), exact inference becomes intractable with increasing evidence (Heckerman, 1989). Given the intractability of exact inference on large, complex networks, researchers have pursued general-purpose, approximate methods based on stochastic sampling such as likelihood weighting (Shachter & Peot, 1989) and Markov chain Monte Carlo simulations (Pearl, 1987). Unfortunately, when applied to large, complex networks such as QMR-BN, these methods also do not scale well (Shwe & Cooper, 1991). In fact, the computational complexity of approximate inference is also known to be NP-hard (Dagum & Luby, 1991).

Even though general-purpose inference algorithms are intractable for large, multiply-connected belief networks, by trading-off generality for time, efficient methods can be found for important special classes of belief networks. Recent work has shown that search-based methods can work well on special classes of large, complex networks (Henrion, 1991; Poole, 1993).

The basic idea behind such search-based inference algorithms for discrete belief networks is to search for high probability partial or complete instantiations of the posterior joint space and then to use these instantiations to derive an estimate of the posterior probabilities of interest.

In general, the success of these search-based methods depends on two factors. The first factor is the skewness of the joint probability mass distribution so that most of the probability is concentrated in a small fraction of the hypotheses. The second factor is the existence of efficient, admissible pruning rules which eliminate large parts of the search space.

The first factor, theoretically and experimentally, appears to be a virtually universal property of belief networks in diagnostic domains (Druzdel, 1994). Skewness commonly arises because most faults or diseases have small prior probabilities, and so, a posteriori, the most probable hypotheses include only one or very few faults.

The second factor, however, depends on the class of belief network. One example is Henrion's TopN algorithm for inference on two-level, noisy-OR belief networks (BN2O) (Henrion, 1991). It uses a powerful pruning rule applicable to two-level networks exhibiting negative product synergy.

In this paper, we describe the TopEpsilon algorithm, a generalization of the TopN algorithm that can handle arbitrary (multi-level), noisy-OR belief networks (NOBNs). TopEpsilon works by efficiently enumerating all complete instantiations of a given belief network consistent with the evidence that have a joint probability $\geq \varepsilon$. These instantiations are then used to compute estimates of posteriors of interest.

For the purposes of algorithm development, we have used a version of the CPCS-BN (Pradhan et al., 1996; Pradhan et al., 1994) in which all nodes are binary and all conditional probability tables are decomposed into noisy-ORs. We call this network CPCS-NOBN. CPCS-BN is a belief-network reformulation of CPCS, a rich, multi-level knowledge base for hepatobiliary disease (Parker & Miller, 1987).

This paper is divided into several parts. In Section 2, we review noisy-OR belief networks, what they are, what some of their properties are, and what algorithms have been developed for them. In Section 3, we briefly review

TopN and point out its strengths and weaknesses. In Section 4, we examine how TopN can be modified for use as a subroutine within TopEpsilon. In Section 5, we look at the TopEpsilon algorithm in detail. In Section 6 we show preliminary experimental results. Finally, in Section 7, we discuss the implications of our work and future plans.

## 2 NOISY-OR BELIEF NETWORKS

Noisy-OR belief networks are belief networks in which influences are modeled as noisy-ORs. The noisy-OR has many important properties, including linearity, factorability, and negative product synergy.

Linearity of the noisy-OR refers to the fact that when using the noisy-OR to model the influence of $k$ parents on a binary node, only $k$ parameters are needed to specify the conditional probability distribution. In contrast, a general influence requires $2^k$ parameters. The linearity of the noisy-OR has been used to reduce the data requirements inherent in building large networks. For example, employing the noisy-OR decomposition of the conditional probability tables (CPTs) in the binary version of CPCS-BN has resulted in an exponential reduction in the number of probabilities that needed to be assessed (Pradhan et al., 1994).

The noisy-OR decomposition of the CPTs can also benefit inference. Heckerman has shown that factoring noisy-OR nodes topologically can reduce the time required for exact inference on noisy-OR belief networks on average by a factor of two to three (Heckerman & Breese, 1994). Zhang has described how algebraic factoring of the noisy-OR in CPCS-NOBN can lead to quick results in some cases (Zhang, 1994). Unfortunately, the two-to-three factor improvement in efficiency exhibited by Heckerman's approach is not enough to make inference on CPCS-NOBN tractable. Zhang's method becomes intractable when non-specific evidence is present.

The last important property is negative product synergy (NPS), which exists between two parents, A and B, on their common successor, C, iff

P(C | A,B) P(C | ~A,~B) < P(C | A,~B) P(C | ~A,B).

The noisy-OR influence exhibits NPS among all the parents of the successor node. The TopN algorithm, described in the next section, exploits this property.

## 3 TOPN

TopN is a search-based algorithm originally developed for inference on two-level, noisy-OR belief networks (BN2O) (Henrion, 1990). It was applied to a BN2O network called QMR-BN, a belief network reformulation of the QMR (Quick Medical Reference) knowledge base (Shwe et al., 1991). BN2O networks such as the one in Figure 1 are two-level networks consisting of a set of diseases, assumed marginally independent, and a set of findings, assumed conditionally independent given any set of diseases (i.e. some diseases present, the rest absent), and noisy-OR influences of diseases on findings. Later TopN was generalized to handle two-level belief networks that exhibit negative product synergy (BN2NPS) (Henrion, 1991).

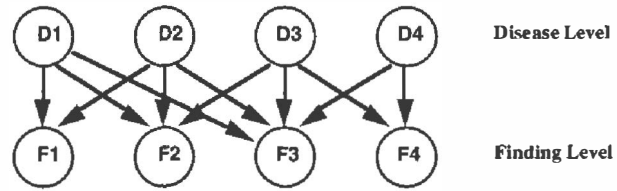

Figure 1: Two-level Network

When applied to the QMR-BN, TopN works effectively on the test cases presented to it. In all twelve test cases, the best estimate of the posteriors converged rapidly toward the correct posteriors.

One weakness of TopN is that it used a best-first search strategy, which requires storage of a potentially exponential list of candidate hypotheses with consequent high demand for memory. As a result, in some of the test cases, the bounds did not converge to the desired precision before running out of memory. What is surprising, however, is that the best-guess estimate of the posterior marginals converged very rapidly in all cases. Another weakness is that TopN is limited to BN2NPS networks.

## 4 TOPN REVISITED

In TopN we have an efficient method for finding the most probable a posteriori instantiations of the disease level of a two-level noisy-OR belief network. Unfortunately, this efficiency comes at the price of space complexity. This problem is easily remedied by making TopN use a depth-first search strategy. We address the second problem, the inapplicability of TopN to multi-level networks, by making a number of modifications to TopN, which involve redefining TopN's output and showing how TopN can be used for another purpose, namely enumerating maximum likelihood (ML) instantiations of the disease level.

### 4.1 TOPN'S OUTPUT REDEFINED

TopN, as its name implies, returns the $N$ most probable instantiations of a two-level network. For the purposes of the TopEpsilon algorithm, we modify TopN to return all those instantiations with a "high" joint probability, where "high" is operationally defined as being a probability $\geq \varepsilon$. This is accomplished by having TopN keep all complete instantiations with probability $\geq \varepsilon$ rather than keeping all instantiations with probability greater than the $N$th best complete instantiation found so far.

### 4.2 ANOTHER USE FOR TOPN

We can use TopN to enumerate ML instantiations of the disease level. That is, we can get TopN to enumerate the instantiations of the disease level that have a high likelihood of causing the evidence in the finding level.

We do this by putting a "wrapper" around TopN. Recall that TopN returns the $N$ most probable a posteriori instantiations of the disease level. Specifically, since P(F), the probability of the observed findings, is not readily available, TopN looks for the $N$ assignments to the disease level nodes that maximize the expression P(D,F). To obtain the top $N$ assignments to the disease level that maximize P(F|D), all we need to do is give TopN uniform priors on nodes in the disease level. Since

$$P(D,F) = P(F|D) P(D) = P(F|D) * \text{constant},$$

TopN in effect returns the $N$ instantiations of the disease level that maximize P(F|D).

Thus, by wrapping TopN with dummy uniform priors, we can make it produce maximum likelihood instantiations of the disease level. Note that if some of the disease nodes in the disease level are observed to be present or absent, then their priors are left unchanged. The wrapped version of TopN will then return the ML instantiations of the unassigned nodes in the disease level. The soundness of the wrapped TopN follows from the soundness of TopN.

### 4.3 EPSILONML

The two modifications to TopN outlined above are independent and can be combined to yield an algorithm called EpsilonML, which returns those instantiations of the disease level that have a likelihood $\geq \varepsilon$ of causing the observed findings in the finding level.

Notice that unlike the original TopN, EpsilonML does not require marginal independence of the disease level nodes. That is, even if the disease level nodes are dependent—that is, arcs exist among the disease level nodes—we can use EpsilonML to enumerate the high likelihood instantiations of the disease level. EpsilonML, however, is still not applicable to networks in which there are dependencies among the nodes in the finding level.

In short, we can use EpsilonML to find the high likelihood instantiations of the immediate parents of a set of evidence among which there are no arcs.

## 5 TOPEPSILON

In this section, we show how EpsilonML can be used to come up with an algorithm for enumerating high probability complete instantiations of a multi-level, noisy-OR belief network. But first, we define what we mean by multi-level.

### 5.1 MULTI-LEVEL NETWORKS

For a network such as the one in Figure 1, it is clear that there are two distinct levels. For a larger network, the number of levels present is not always so clear. To clear-up any ambiguity, we define a node's level to be the greatest number of arcs between it and a root node, where a root node is a node with no parents.

```
LabelNodes(belief-network)

    current-level := 0
    parent-node-list := root nodes in
belief-network
    FOR each node in parent-node-list DO
        level(node) := current-level
    REPEAT
        successor-list := union of child
nodes of the nodes in parent-node-list
        current-level := current-level + 1
        FOR each node in successor-list DO
            level(node) := current-level
        parent-node-list := successor-list
    UNTIL successor-list is empty
    RETURN current-level
```

Figure 2: Pseudocode for LabelNodes(belief-network) which labels each in node in the belief-network by its level and returns the maximum level.

Node-level labeling can be accomplished by performing a breadth-first traversal of a belief network starting at the root nodes. The pseudocode is in Figure 2. For example, consider the simple belief network in Figure 3. The root nodes are labeled with level 0. The immediate children of the root nodes are labeled with level 1. Then their children are labeled with level 2. Notice that node E is actually labeled twice. It is first a level 1 node because it a child of the root node A. But because it is also a child of node C, a level 1 node, it is assigned to level 2. The breadth-first labeling guarantees that a node keeps the label it got last, which is its farthest "distance" from a root node.

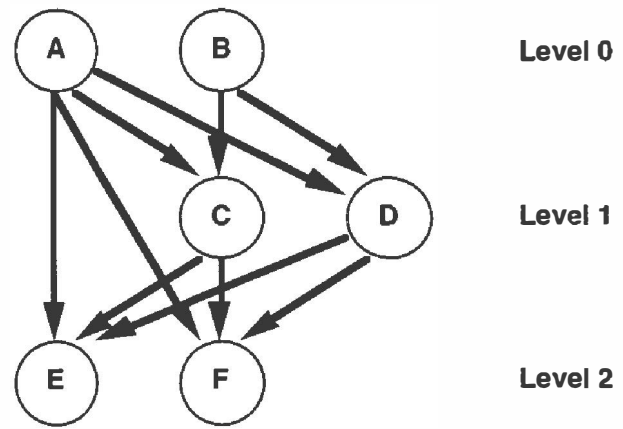

Figure 3: Example of Level Labeling

The number of levels a belief network has is one plus the maximum level label any node in the network receives. Thus, the network in Figure 3 is a three-level network.

The importance of labeling the nodes of a network by their level is that it provides us a way of breaking a multi-level network into a series of two-level networks, each of which is a subproblem for EpsilonML.

### 5.2 BASIC IDEA

Starting with the evidence nodes at the deepest level, TopEpsilon constructs a complete instantiation incrementally, level-by-level, using EpsilonML. The $\varepsilon$ fed to EpsilonML changes dynamically in accordance with the likelihood of the partial instantiation so far and $\varepsilon_{target}$ for the full network.

We illustrate the basic idea using a simple example.

### 5.3 EXAMPLE

Consider the network in Figure 4. Clearly, it is a three-level network.

Let the evidence be that C=c. Assume that we want to enumerate all complete instantiations of the network that have a joint probability that is $\geq \varepsilon_{target}$.

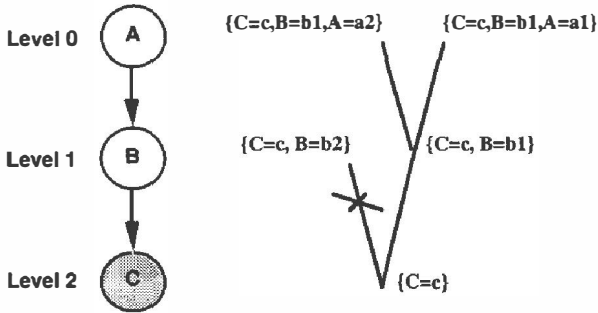

Figure 4: Example Network and Search Tree

Starting with the evidence nodes in the deepest level, node C in level 2, we look for high likelihood instantiations of its immediate parents. In this case, the evidence at level 2 has only one parent, namely node B. To find an assignment to node B that has a high likelihood of causing C=c, we call EpsilonML with an $\varepsilon$ equal to $\varepsilon_{target}$: EpsilonML($\varepsilon_{target}$). EpsilonML returns a list of instantiations of B that satisfy the requirements. In this case, suppose that only B=b1 is returned, meaning that B=b2 is not likely to cause C=c. At this point we have a partial instantiation of the three-level network that has a likelihood $\geq \varepsilon_{target}$. The next step is to call EpsilonML again to find high likelihood assignments to node A given that C=c and B=b1. This time, however, we send a different $\varepsilon$ to EpsilonML. The $\varepsilon$ we send it must take into account the likelihood of the current partial instantiation. Specifically,

$$\varepsilon_{new} = \varepsilon_{target}/P(C=c|B=b1).$$

This comes from the simple observation that if we want $P(A,B,C) \geq \varepsilon_{target}$ and we know $P(C|B)$, then we must have

$$P(C=c|B=b1)P(B=b1|A)P(A) \geq \varepsilon_{target},$$

which is equivalent to

$$P(B=b1|A)P(A) \geq \varepsilon_{target}/P(C=c|B=b1).$$

Since $P(A) \leq 1$, we must have

$$P(B=b1|A) \geq \varepsilon_{target}/P(C=c|B=b1).$$

In general, the $\varepsilon_{new}$ is equal to the ratio of $\varepsilon_{target}$ to the likelihood of the partial instantiation so far. In the pseudocode, the likelihood of the partial instantiation is computed using the function probability(), which is defined as the product of the known terms in the factorization of the joint probability implied by the belief network structure.

### 5.4 THE ALGORITHM

TopEpsilon generalizes the approach taken in the above example. The pseudocode is given below:

```
TopEpsilon(εtarget, belief-network)
    LabelNodes (belief-network)
    max-level := level of the deepest evidence
        nodes
    FOR each node in belief-network DO
        score(node) := 0
    mass-accumulated := 0
    initialize stack to empty
    current-state := partial assignment con-
        sisting of evidence at max-level
    PUSH current-state onto stack
    WHILE stack is not empty DO
        BEGIN WHILE
            current-state := pop(stack)
            IF complete(current-state) AND
    probability(current-state) ≥ εtarget
                THEN
                    mass-accumulated := mass-
    accumulated + probability(current-
    state)
                ELSE
                    εnew := εtarget/probability(cur-
    rent-state)
                    extension-list := EpsilonML(cur-
    rent-state, εnew)
                    push extension-list onto stack
            END ELSE
        END WHILE
    FOR each node in belief-network DO
        estimated-posterior(node) :=
        score(node)/mass-accumulated
```

## 6 EXPERIMENTS

We have done some preliminary experiments to examine the performance of TopEpsilon. In particular, we have measured TopEpsilon's resource use (time complexity as a function of $\varepsilon$) and convergence properties (fraction of total mass accumulated as a function of time, $\varepsilon$, and amount of evidence).

We ran TopEpsilon on the BN3 network, a five-level sub-network of the CPCS-NOBN containing 3 diseases, 97 findings, and 146 nodes total. We used a set of cases generated by sampling from the CPCS-NOBN. Each case contains from zero to three diseases and 26 or 83 findings.

Given an $\varepsilon_{target}$, TopEpsilon will search for all complete instantiations with probability $\geq \varepsilon_{target}$. If $\varepsilon_{target}$ is too large, say 0.1, then TopEpsilon may return no plausible explanations for the evidence. If $\varepsilon_{target}$ is too small, then TopEpsilon may accumulate more joints than it really needs to give an answer with some desired precision. Thus, we need a way to choose the right $\varepsilon_{target}$ for the precision desired. We have no simple way of doing this right now. Therefore, for the preliminary experiments, we used a fixed $\varepsilon$ schedule, starting at $\varepsilon_{target} = 1.0E-2$ and reducing by powers of 100 to 1.0E-20. In the future, we plan to wrap an adaptive $\varepsilon$ scheduler around TopEpsilon.

For a gold standard, $\varepsilon_{target}$ was set to zero. In this way, we compute the total probability mass consistent with the evidence in each test case.

TopEpsilon is implemented in Macintosh Common Lisp 3.0 and ran on a PowerBook 5300cs using Speed Doubler and 15MB of allocated memory.

### 6.1 RESOURCE USE

As we lower $\varepsilon$, we expect the state space will increase in size. Without pruning, the space of possible extensions will grow exponentially. With pruning, the hope is that the combinatorial explosion can be avoided. In the graph below, where each line corresponds to a different test case containing 26 findings, we show that the number of partial instantiations (states) explored as a function of $-\log(epsilon)$ is subexponential. It turns out that because the time complexity is approximately linear in the number of states explored, that the time complexity as a function of $-\log(epsilon)$ is also subexponential. (Space complexity, because of TopEpsilon's modified depth-first search strategy, is approximately linear in the depth of the search tree, which never exceeds 4 for BN3.)

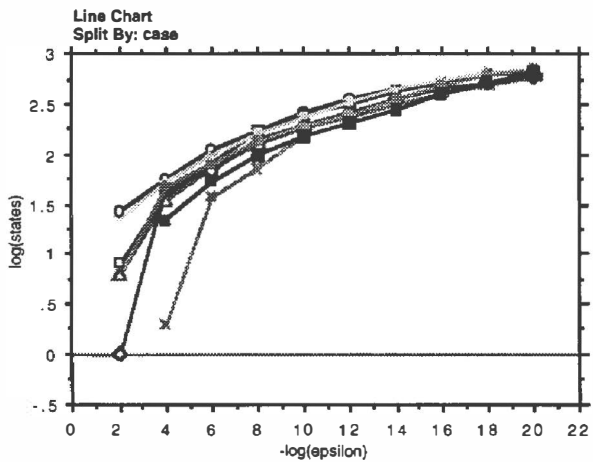

### 6.2 CONVERGENCE

#### 6.2.1 Convergence as a Function of Time

The lower the $\varepsilon$, the longer TopEpsilon takes to run. In the graph below, we show that for test cases containing 26 findings, the mass accumulated by TopEpsilon converges rapidly to the gold standard.

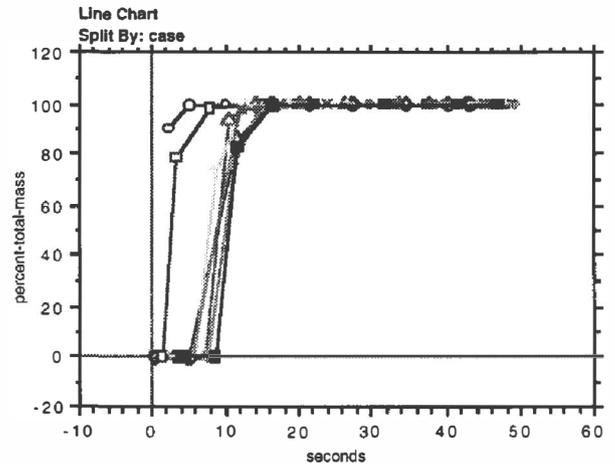

#### 6.2.2 Convergence as a Function of Epsilon

The prior probability of evidence in each test case varies with the number of diseases. Thus, we expect that different cases will start accumulating mass at different $\varepsilon$'s. The graph below shows that in the test cases with 26 findings that once mass begins to be accumulated, convergence is rapid.

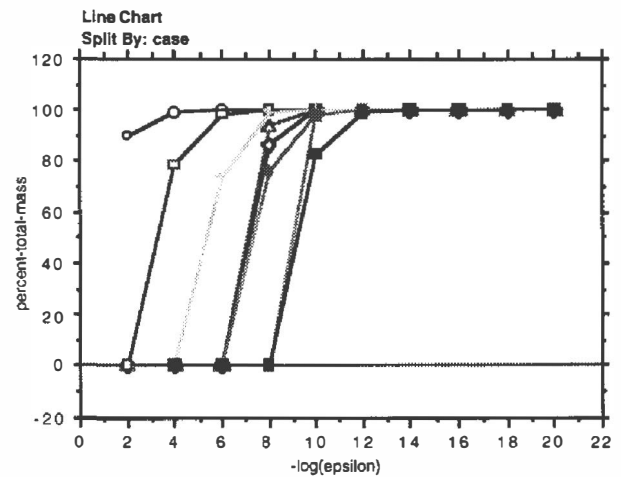

#### 6.2.3 Convergence as a Function of Evidence

Below is a graph identical to the one in Section 6.2.2, except that for each test case, more findings are available, resulting in a total of 83 findings for each test case. We see that the presence of more evidence causes the curves to shift to the right relative to the curves in the graph in 6.2.2. This shift is not surprising and corresponds to a decrease in prior probability of the evidence. What is striking, however, is that the span of $\varepsilon$'s over which most of the mass in a given case is accumulated is roughly unchanged.

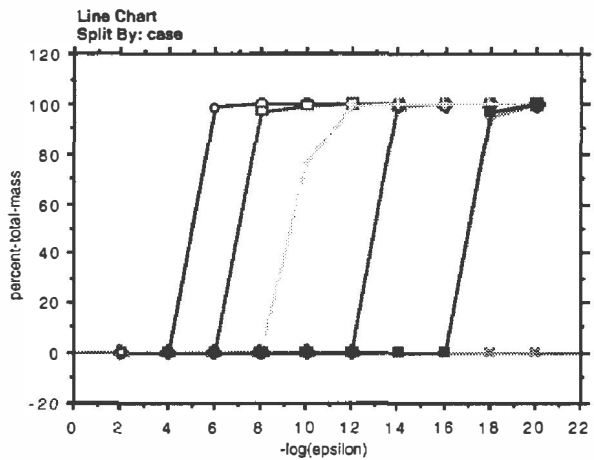

## 7 DISCUSSION AND FUTURE PLANS

The experimental results are encouraging and suggest that the simple use of TopN as an extension generator for multi-level networks can be very effective. However, more work needs to be done. In particular, bounds need to be developed for the estimated posteriors output by the algorithm. Bounds analogous to those used by the TopN algorithm could clearly be used. However, to avoid slow convergence of the absolute bounds, better bounding techniques will likely need to be employed.

Strictly speaking, the depth-first scheme we have used to explore the space of partial instantiations is not linear in space complexity. The modified version of TopN being used as a branching operator may still cause us to run out of memory while looking for admissible one-level extensions of a given partial instantiation. Implementation of a true depth-first search strategy would remedy this problem. However, given the small size of our test network, we have not observed a memory bound.

Finally, as currently implemented, prior information, unless present in a two-level subproblem, is not utilized. Thus, the search is largely data-driven. In cases where a lot of evidence is available, it makes sense to take a data-driven, maximum-likelihood approach. In cases where the priors are dominant and evidence is so nonspecific as to not cause the posterior distribution to deviate much from the prior distribution, the approach taken here will likely suffer. The presence of more evidence constrains the search. Less evidence constrains the search less, leading to inefficiency. More experimental work needs to be done to characterize how TopEpsilon scales as a function of network depth and breadth.

## Acknowledgments

This work was supported by NLM grant LM07033, NSF Grant Project IRI-9120330, and by computing resources provided by the Stanford University CAMIS project, which is funded under grant number LM05305 from the National Library of Medicine. The authors would also like to thank R. Parker, R. Miller, and the University of Pittsburgh for access to the CPCS knowledge base.